\documentclass[runningheads]{llncs}
\usepackage[T1]{fontenc}
\usepackage{amsmath}
\usepackage{graphicx}
\usepackage{amsmath}  
\usepackage{amssymb}  

\usepackage{graphicx} 
\usepackage{hyperref}
\hypersetup{
    colorlinks=true,
    linkcolor=blue,    
    citecolor=blue,     
    filecolor=magenta, 
    urlcolor=cyan      
}
\usepackage{subcaption} 
\usepackage{tikz}
\def\checkmark{\tikz\fill[scale=0.4](0,.35) -- (.25,0) -- (1,.7) -- (.25,.15) -- cycle;} 

\begin{document}
\title{IDAL: Improved Domain Adaptive Learning for Natural Images Dataset}
%

%
\author{Ravi Kant Gupta \and
Shounak Das \and Amit Sethi}
%
%
\institute{Indian Institute of Technology Bombay, Powai, Mumbai, India\\
\email{\{ravigupta131,21D070068,asethi\}@iitb.ac.in}}


%
\maketitle              
\begin{abstract}
We present a novel approach for unsupervised domain adaptation (UDA) for natural images. A commonly-used objective for UDA schemes is to enhance domain alignment in representation space even if there is a domain shift in the input space. Existing adversarial domain adaptation methods may not effectively align different domains of multi-modal distributions associated with classification problems. Our approach has two main features. Firstly, its neural architecture uses the deep structure of ResNet and the effective separation of scales of feature pyramidal network (FPN) to work with both content and style features. Secondly, it uses a combination of a novel loss function and judiciously selected existing loss functions to train the network architecture. This tailored combination is designed to address challenges inherent to natural images, such as scale, noise, and style shifts, that occur on top of a multi-modal (multi-class) distribution. 
The combined loss function not only enhances model accuracy and robustness on the target domain but also speeds up training convergence. 
Our proposed UDA scheme generalizes better than state-of-the-art for CNN-based methods on Office-Home, Office-31, and VisDA-2017 datasets and comaparable for DomainNet dataset.
\end{abstract}
\section{Introduction}

Unsupervised Domain Adaptation (UDA) addresses the performance degradation caused by domain shift in supervised learning, where there's a significant distribution difference between training (source) and testing (target) data domains. Adversarial-based UDA, utilizing frameworks like Generative Adversarial Networks (GANs)~\cite{pavanetto2021generation} and Domain Adversarial Neural Networks (DANN)~\cite{ganin2016domain}, aims to mitigate this by learning domain-invariant features from unlabeled target data. By promoting feature harmonization while retaining class information, these models enhance target domain generalization. Despite promising results in image classification and object detection, adversarial UDA faces challenges such as hyper-parameter sensitivity, high-dimensional space navigation, and domain shift detection. 

To address the aforementioned challenge, we developed an unsupervised domain adaptation approach that surpasses the state-of-the-art UDA performance of convolution neural networks (CNNs) for benchmark natural image datasets -- Office-Home~\cite{venkateswara2017deep}, Office-31~\cite{saenko2010adapting}, VisDA-2017~\cite{pei2018multi}, and shows comparable results for DomainNet~\cite{peng2019moment}.

Inspired by the concept of a conditional domain adversarial network (CDAN) ~\cite{long2018conditional}, our core approach -- Improved Domain Adaptive Learning (IDAL) -- involves concurrent training of a feature extractor (typically a deep neural network) and a domain classifier (discriminator) tasked with distinguishing between source and target domains. We have explored various CNN-based feature extractors such as ResNet-101, ResNet-50~\cite{jian2016deep}, ViT~\cite{dosovitskiy2020image}, and ConvMixer~\cite{trockman2022patches} to extract meaningful features. The feature extractor's aim is to learn representations that are invariant to domain shifts, and thus deceive the domain classifier that endeavors to correctly classify the domain of the extracted features. The integration of ResNet-50 and FPN combines~\cite{lin2017feature} deep feature representation and multi-scale extraction, essential for tasks like object detection and segmentation. Given that object scale and style vary by domain, this synergy makes ResNet + FPN a strong candidate for Unsupervised Domain Adaptation (UDA), focusing on higher-level domain-specific feature suppression. This application to UDA is novel.

In the adversarial training process, the feature extractor (ResNet-50 + FPN) and domain classifier compete: the extractor aims to produce domain-agnostic features, while the classifier attempts to distinguish between domains. This method fosters the development of domain-invariant features, enhancing transferability across source and target domains.

To improve the training process, we propose a novel loss function called pseudo label maximum mean discrepancy (PLMMD). We use this loss in addition to certain existing losses -- maximum information loss (entropy loss)~\cite{krause2010discriminative}, 
maximum mean discrepancy (MMD) loss~\cite{long2015learning}, 
minimum class confusion (MCC) loss~\cite{jin2020minimum}. Our model integrates several loss functions to enhance domain adaptation and classification accuracy: Maximum information loss clusters target features by class, preserving key information. MMD loss bridges domain gaps by comparing mean embeddings. MCC loss boosts accuracy by minimizing class confusion, vital for uneven datasets. Our innovative PLMMD loss selectively extracts domain-invariant features, speeding up training. This tailored mix of loss functions enables our method to outperform existing CNN models and achieve quicker convergence on natural image datasets-- Office-Home~\cite{venkateswara2017deep}, Office-31~\cite{saenko2010adapting}, and  VisDA~\cite{pei2018multi}.

\section{Background and Related Works}

In unsupervised domain adaptation (UDA), we have data from a source domain $D_s = \{(x_{s_i}, y_{s_i})\}_{i=1}^{n_s}$ as ${n_s}$ labeled examples and that from a target domain $D_s = \{(x_{t_i}, y_{t_i})\}_{i=1}^{n_t}$ as ${n_t}$ unlabeled examples where $y_{t_i}$'s are unknown. The source domain and target domain are sampled from the distributions $P({x_s},{y_s})$ and $Q({x_t},{y_t})$ respectively. Notably, the two distributions are initially not aligned; that is, $P \neq Q$. 

Domain adversarial neural network (DANN)~\cite{ganin2016domain} is a framework of choice for UDA. This is a dual-player game involving two key components: the domain discriminator, denoted as $D$, and the feature representation, denoted as $F$. In this setup, $D$ is trained to differentiate between the source domain and the target domain, while $F$ is simultaneously trained to both confound the domain discriminator $D$ and accurately classify samples from the source domain.  The discrepancy between the feature distributions $P_F$ and $Q_F$~\cite{ganin2015unsupervised} has well corresponding with the error function of the domain discriminator. This is a key to bound the risk associated with the target in the domain adaptation theory~\cite{ben2010theory}.


An alternative approach in the field of Unsupervised Domain Adaptation (UDA) focuses on reducing the domain discrepancy as quantified by various metrics, e.g., maximum mean discrepancy(MMD).
To establish class-level alignment across domains, the methodology outlined in the study conducted by Pei and colleagues~\cite{pei2018multi} incorporates a multiplicative interaction between feature representations and class predictions. In their studies~\cite{chen2019progressive}, efforts are made to ensure alignment between the centroids of labeled source data and the centroids derived from pseudo-labeled target data, particularly for shared classes within the feature space.

Another approach to UDA involves employing separate task classifiers for each of the two domains. These classifiers are used to identify non-discriminative features. In turn, they facilitate the learning of a feature extractor that focuses on generating discriminative features~\cite{lee2019sliced}. Several other studies emphasize the importance of directing attention towards transferable regions as a means to establish a domain-invariant classification model, as exemplified by~\cite{kurmi2019attending}. In addition, for the purpose of extracting target-discriminative features,\cite{kang2018deep} employ techniques such as generating synthetic data from the raw input of the two domains, as described in\cite{pavanetto2021generation}.




Since our work modifies the network and losses of the CDAN framework~\cite{long2018conditional}, we explain it here for completeness. To reduce the shift in data distributions across the domains, CDAN trains a deep network $N: x \rightarrow y$, so that source risk $r_s$= \(E_{(x_s, y_s) \sim P}[N(x_s) \neq y_s]\) can bound the target risk $r_t$= \(E_{(x_t, y_t) \sim Q}[N(x_t) \neq y_t]\) plus the distribution discrepancy $disc(P, Q)$ quantified by a novel conditional domain discriminator. In the context of adversarial learning, Generative Adversarial Networks (GANs)~\cite{ganin2016domain} play a pivotal role in mitigating differences between domains. A deep network $N$ generates features represented by $f = F (x)$ and classifier prediction denoted by $g = N(F(x))$.

We enhance existing methods for adversarial domain adaptation in two specific ways. Firstly, when dealing with non-identical joint distributions of features and classes across domains, as characterized by $P({x_s},{y_s})$ and $Q({x_t},{y_t})$, relying solely on the adaptation of the feature representation $f$ may prove insufficient, as highlighted in~\cite{long2018conditional}. A quantitative analysis indicates that deep representations tend to transition from a more general to a domain-specific nature as they traverse deeper layers within neural networks. This transition leads to a notable decrease in transferability, particularly observed in the layers responsible for domain-specific feature extraction ($f$) and classification ($g$), as detailed in~\cite{yosinski2014transferable}. Secondly, due to the nature of multi-class classification, the feature distribution is multi-modal, and hence adapting feature distribution may be challenging for adversarial networks.

Simultaneous modeling the domain variances in feature representation $(f)$ and classifier prediction $(g)$ facilitates effective domain gap reduction~\cite{long2018conditional}. This joint conditioning helps capture and align data distributions between source and target domains. Thus, incorporating classifier prediction as a conditioning factor in domain adaptation shows promising potential for enhancing transferability and producing domain-invariant representations in challenging cross-domain scenarios. CDAN originally introduced a minimax optimization framework featuring two adversarial loss terms: (a) the source classifier loss, aimed at minimizing it to ensure a lower source risk, and (b) the discriminator loss applied to both the source classifier $N$ and the domain discriminator $D$, spanning both the source and target domains. This loss is minimized with respect to $D$ while simultaneously maximized with respect to $f = F(x)$ and $g = N(F(x))$:
\begin{equation}
  L_{\text{clc}}(x_{s_i}, y_{s_i}) = \mathbb{E}_{(x_{s_i}, y_{s_i}) \sim D_s} \, L(N(x_{s_i}), y_{s_i})
  \label{eqn1}
\end{equation}

\begin{equation}
\begin{aligned}
L_{dis}(x_s, x_t) = & -\mathbb{E}_{x_{s_i} \sim D_s} \log [D(f_{s_i}, g_{s_i})] \\
         & -\mathbb{E}_{x_{t_j} \sim D_t} \log [1 - D(f_{t_j}, g_{t_j})],
\end{aligned}
\label{eqn2}
\end{equation}
In this context, $L$ corresponds to the cross-entropy loss, $L_{clc}$ is the classifier loss, $L_{dis}$ is the discriminator loss and $h = (f, g)$ signifies the combined variable encompassing the feature representation $f$ and classifier prediction $g$. The minimax game of CDAN is
\begin{equation}
\begin{aligned}
\begin{gathered}
\min_{N} L_{clc}(x_{s_i}, y_{s_i}) - \lambda L_{dis}(x_s, x_t) \\
\min_{D} L_{dis}(x_s, x_t),
\end{gathered}
\end{aligned}
\label{eqn3}
\end{equation}
Here, $\lambda$ denotes a hyper-parameter that balances between the two objectives, allowing for a trade-off between source risk and domain adversary concerns.

\begin{figure}[!]
\centering
\includegraphics[height=6cm,width=8cm]{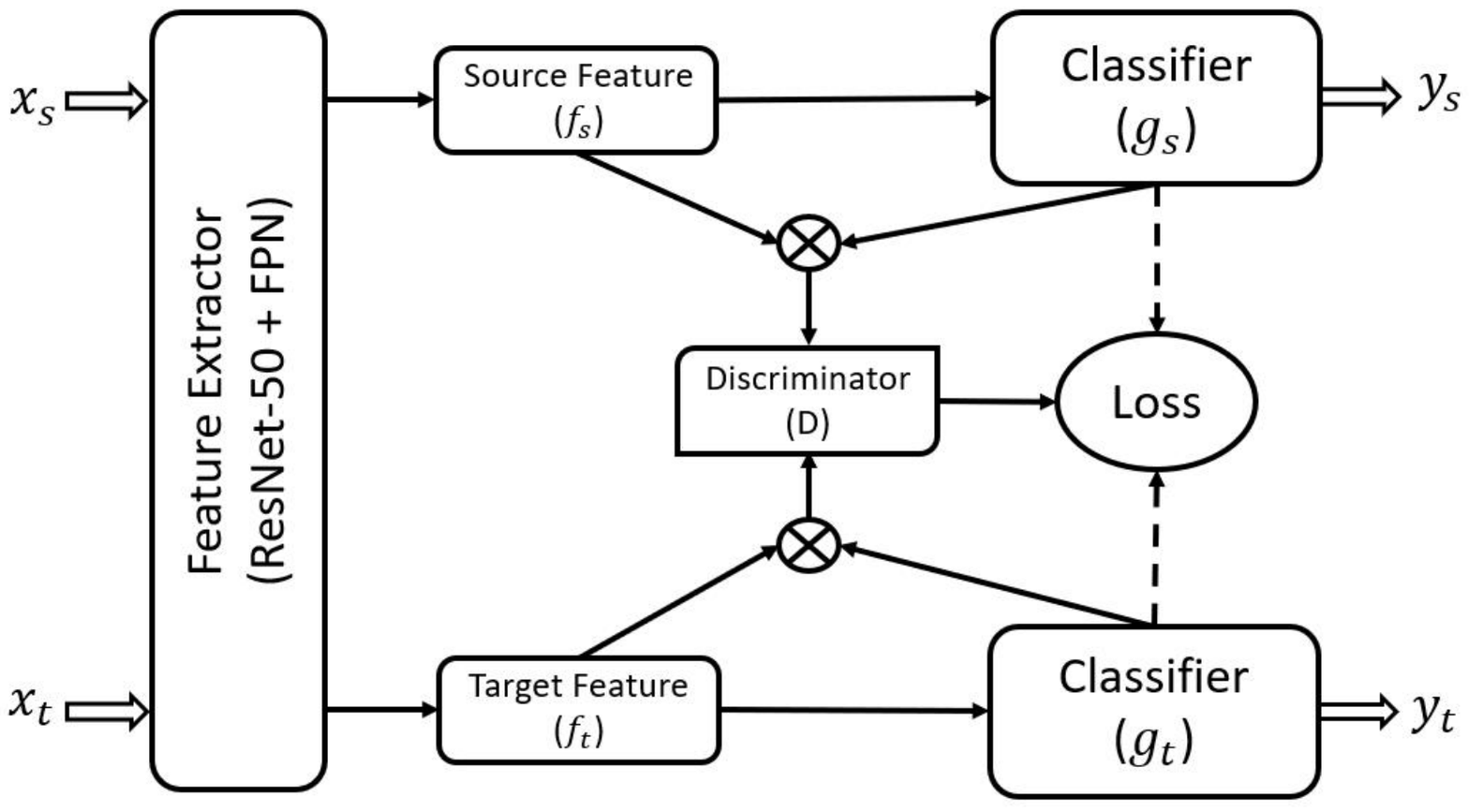}
\vspace{-8mm}
\caption{The CDAN~\cite{long2018conditional} architectural framework, illustrating the joint reduction of the cross-domain gap by the conditional domain discriminator $D$, with domain-specific feature representation $f$ and classifier prediction $g$ at its core. The symbol $\otimes$ signifies a multilinear mapping operation.}
\label{fig2}
\end{figure}

As depicted in Figure~\ref{fig2}, the domain discriminator $D$ is conditioned on the classifier prediction $g$ via the joint variable $h = (f, g)$, aiming to address the two challenges inherent in adversarial domain adaptation, as discussed in~\cite{long2018conditional}. To incorporate a basic form of conditioning for $D$, we employ $(D(f \oplus g))$—where we concatenate the feature representation and classifier prediction into a vector $(f \oplus g)$, which serves as the input for the conditional domain discriminator $D$. This conditioning approach aligns with the common practice observed in existing conditional GANs~\cite{ganin2016domain}. However, when utilizing the concatenation approach, $f$ and $g$ operate independently, missing the opportunity to fully capture the crucial multiplicative interactions between the feature representation and classifier prediction that play a pivotal role in domain adaptation. A multilinear map is formed by computing the outer product of multiple random vectors. This technique, which involves multilinear maps applied to infinite-dimensional nonlinear feature maps, has proven successful in embedding joint or conditional distributions into reproducing kernel Hilbert spaces~\cite{song2009hilbert}. 
In addition to the theoretical advantages offered by the multilinear map $(x \otimes y)$ in comparison to concatenation $(x \oplus y)$, as discussed in~\cite{song2009hilbert}. In this research, we harness the capabilities of the multilinear map to condition $D$ on $g$. In contrast to concatenation, the multilinear map, denoted as $(x \otimes y)$, excels in capturing the intricate multi-modal structures that are inherent in complex data distributions. However, it's important to note that a drawback of the multilinear map is its potential for dimension explosion.

\vspace{-0.8mm}
Our approach involves the joint minimization of the source classifier $N$ and feature extractor $F$ with respect to Equation (\ref{eqn1}). Additionally, we minimize Equation (\ref{eqn2}) to optimize the domain discriminator $D$ and simultaneously maximize Equation (\ref{eqn2}) to enhance the feature extractor $F$ and source classifier $N$. This yields the mini-max problem of Domain Adversarial Networks:

\begin{align}
\min_{G} \quad & \mathbb{E}_{(x_s^i, y_s^i)\sim D_s} L(G(x_s^i), y_s^i) \notag \\
& + \lambda\left( \mathbb{E}_{x_s^i \sim D_s}\log [D(T(h_s^i))] \right. \notag \\
& \left. + \mathbb{E}_{x_t^j \sim D_t}\log[1 - D(T(h_t^j))]\right) \notag\\
\max_{D} \quad & \mathbb{E}_{x_s^i \sim D_s} \log [D(T(h_s^i))] + \mathbb{E}_{x_t^j \sim D_t} \log [1 - D(T(h_t^j))], 
\end{align}
In this context, $\lambda$ serves as a hyper-parameter responsible for adjusting the weightage between the source classifier and the conditional domain discriminator and $G$ acts as the generator. Meanwhile, $h = (f, g)$ represents the composite variable encompassing both the domain-specific representation $f$ and the classifier prediction $g$ which play pivotal roles in adversarial adaptation.
\cite{gupta2023domain} shows the improvement in performance with use of tailored loss function for medical datasets. 

\section{Proposed Method}

In our Unsupervised Domain Adaptation (UDA) strategy for cross-domain classification, we leverage insights from a labeled source domain to enhance target domain performance, despite unlabeled conditions. Our method utilizes a multi-scale optimized neural architecture, ensuring well-separated, multi-modal class distributions. Data augmentation techniques such as flipping, resizing, and normalization are applied for domain consistency. A novel aspect of our approach is a unique loss function that, combined with selected existing ones, minimizes domain discrepancies and aligns feature distributions across multi-class datasets with varying image sizes. For augmented image feature extraction, we employ ResNet-50 coupled with the Feature Pyramid Network (FPN), blending deep feature capture with multi-scale extraction to effectively represent detailed and broad image features, a pioneering application in UDA and image classification.

\subsection{Proposed loss function}

The proposed loss function to train the new architecture in the CDAN~\cite{long2018conditional} framework for improving UDA for image classification can be formulated as follows:

\begin{equation}
\begin{aligned}
L=\min_{N} L_{clc}(x_{s_i}, y_{s_i}) - \lambda L_{dis}(x_s, x_t)\\ +\beta L_{IM} + \gamma L_{MCC} + \delta L_{MDD} +  \eta L_{PLMMD},
\end{aligned}
\end{equation}
where $\lambda$, $\beta$, $\gamma$, $\delta$ and $\eta$ are hyper parameters, $L_{IM}$  is the information maximization (entropy) loss, $L_{MCC}$ is minimum class confusion loss, $L_{MDD}$ is maximum mean discrepancy loss, and $L_{PLMDD}$ is a novel pseudo-label maximum mean discrepancy loss. It is worth noting that the original CDAN~\cite{long2018conditional} trained a ResNet (and not ResNet $+$ FPN as proposed) using only $L_{clc}, L_{dis}$, and $L_{IM}$. On the other hand, all other individual loss terms have their own specialty and this novel combination of loss significantly surpasses the performance of CNN-based as well as transformer-based models. A detailed description of all the losses, including the proposed $L_{PLMDD}$ are given below.
\subsubsection{Information Maximization loss:}
The Information Maximization loss is designed to encourage neural networks to learn more informative representations by maximizing the mutual information between the learned features and the input data~\cite{krause2010discriminative}. 
By maximizing the mutual information between the empirical distribution of target inputs and the resulting distribution of target labels, which can be formally defined as:

\begin{equation}
\begin{gathered}
I(p_t; x_t) = H(\overline{p}_t) - \frac{1}{n_t} \sum_{j=1}^{n_t} H(p_{tj}) \\
= -\sum_{k=1}^{K} \overline{p}_{tk} \log(\overline{p}_{tk}) + \frac{1}{n_t} \sum_{j=1}^{n_t} \sum_{k=1}^{K} p_{tkj} \log(p_{tkj})
\end{gathered}
\end{equation}
where, \(p_{tj} = \text{softmax}(G_c(G_f(x_{tj})))\), \(\overline{p}_t = \mathbb{E}_{x_t}[p_t]\), and K is the number of classes.
By taking into account \(I(p_t; x_t)\), our model is incentivized to learn target features that exhibit tight clustering along with a uniform distribution. This approach is designed to retain discriminative information within the target domain.

\subsubsection{Minimum Class Confusion:} 
The minimum class confusion loss, referenced as $\mathcal{L}_{MCC}$~\cite{jin2020minimum}, aims to mitigate confusion between various classes represented by indices $j$ and $j'$, where these indices collectively encompass the entire set of classes. Notably, this loss term is focused on the target domain and is intended to minimize the confusion between pairs of classes, such as those denoted by $j$ and $j'$ is given by: 
\begin{center}
     $C_{jj'} =  \hat{\mathbf{y}}_{\cdot j}^{\intercal} W\hat{\mathbf{y}}_{\cdot j'}^{\intercal}$ 
\end{center} 
 
After standardizing (normalising) the class confusion terms, the ultimate MCC Loss function is defined as:
 \begin{equation}
     \mathcal{L}_{MCC} = \frac{1}{c} \sum_{j=1}^ {c} \sum_{j' \neq j}^ {c}|C_{jj'}|,
 \end{equation}
This loss is computed as the summation of all non-diagonal elements within the class confusion matrix. The diagonal elements signify the classifier's level of "certainty," whereas the non-diagonal elements signify the "uncertainty" associated with classification. The MCC loss can be incorporated alongside other domain adaptation techniques.

\subsubsection{Maximum Mean Discrepancy:}
Maximum Mean Discrepancy (MMD) is a kernel-based two-sample statistical test employed to assess the similarity between two distributions. 
The final loss for a given probability measure $P$ and $Q$ takes the following form:
\begin{equation}
\begin{aligned}
\begin{gathered}
{L}_{MMD}=MMD^2 (P, Q) \\= \mathbb{E}_{P}[k(X,X)] - 2\mathbb{E}_{P,Q}[k(X,Y)] +\mathbb{E}_{Q}[k(Y,Y)]
\end{gathered}
\end{aligned}
\label{eqn 11}
\end{equation}

\subsubsection{Pseudo-label MMD:}

We propose a novel loss function called pseudo-label maximum mean discrepancy (PLMMD). This loss function takes into account pseudo-labels that can be generated on the target domain samples after the first few training iterations. Doing so strongly conditions the feature alignment on the classes. It is calculated using a procedure similar to that of calculating MMD. The difference is that we multiply each of the expectations in Equation~\ref{eqn 11} with weights that are calculated based on pseudo-labels:
\begin{equation}
\begin{aligned} 
\begin{gathered}
{L}_{PLMMD}= w_{XX} \mathbb{E}_{P}[k(X,X)] - 2 w_{XY} \mathbb{E}_{P,Q}[k(X,Y)] \\+ w_{YY}\mathbb{E}_{Q}[k(Y,Y)],
\end{gathered}
\end{aligned}
\end{equation}
where $w_{XX}$ represents weight to get similarity within the source domain,  $w_{YY}$ are weights for similarity within the target domain, and $w_{XY}$ are weights to get similarity within the source and target domain. For calculating the weights, firstly source and target label data are normalized to account for class imbalances. For each class common to both datasets, dot products of normalized vectors are computed to quantify instance relationships. Calculated dot products are normalized by the count of common classes, ensuring fairness. This returns three weight arrays, representing relationships between instances in the source dataset, target dataset, and source-to-target pairs.

\begin{table*}[b!]
\caption{Comparison with SoTA methods on Office-Home. IDAL(ours) is reported with and without(w/o) FPN. The best performance is marked as bold, and the second best is underlined}\label{t2}
\begin{center}
\resizebox{\linewidth}{!}{
\begin{tabular}{|c|c|c|c|c|c|c|c|c|c|c|c|c|c|}
    \hline
    \textbf{Model} & \textbf{A$\rightarrow$C} & \textbf{A$\rightarrow$P} & \textbf{A$\rightarrow$R} & \textbf{C$\rightarrow$A} & \textbf{C$\rightarrow$P} & \textbf{C$\rightarrow$R} & \textbf{P$\rightarrow$A} & \textbf{P$\rightarrow$C} & \textbf{P$\rightarrow$R} & \textbf{R$\rightarrow$A} & \textbf{R$\rightarrow$C} & \textbf{R$\rightarrow$P} &\textbf{Avg.}\\ \hline \hline
     ResNet-50 \cite{jian2016deep}  & 34.9   &50.0  &58.0  & 37.4  &41.9 &46.2 &38.5  &31.2 &60.4 &53.9 &41.2 &59.9 &46.1  \\ 
     DANN \cite{ganin2016domain} &	45.6	&	59.3	&	70.1	&	47.0	&	58.5	&	60.9	&	46.1	&	43.7	&	68.5	&	63.2	&	51.8	&	76.8	&	57.6	\\ 
     CDAN \cite{long2018conditional}  &	50.7	&	70.6	&	76.0	&	57.6	&	70.0	&	70.0	&	57.4  &	50.9	&	77.3	&	70.9	&	56.7	&	81.6	&	65.8	\\ 
     MDD \cite{pmlr-v97-zhang19i} &	54.9	&	73.7	&	77.8	&	60.0	&	71.4	&	71.8	&	61.2	&	53.6	&	78.1	&	72.5	&	60.2	&	82.3	&	68.1	\\ 
     GVB-GD \cite{cui2020gradually} &	57.0	&	74.7	&	79.8	&	64.6	&	74.1	&	74.6	&	65.2	&	55.1	&	81.0	&	74.6	&	59.7	&	84.3	&	70.4	\\ 
     SRDC \cite{tang2020unsupervised} &	52.3	&	76.3	&	81.0	&	\underline{69.5}	&	76.2	&	78.0	&	68.7	&	53.8	&	81.7	&	76.3	&	57.1	&	85.0	&	71.3	\\
     SHOT \cite{liang2020we} &	56.9	&	\textbf{78.1}	&	\underline{81.0}	&	67.9	&	\underline{78.4}	&	\underline{78.1}	&	67.0	&	54.6	&	81.8	&	73.4	&	58.1	&	84.5	&	71.6	\\ 
     SDAT \cite{rangwani2022closer} &	58.2	&	77.1	&	\textbf{82.2}	&	66.3	& 77.6 &	76.8	&	63.3	&	57.0	&82.2	&	74.9	&	\textbf{64.7}	&	\underline{86.0}	&	72.2	\\ 
     FixBi \cite{na2021fixbi} & 58.1   &77.3  &80.4  & 67.7  &\textbf{79.5} &\textbf{78.1} &65.8  &\textbf{57.9}&81.7 &76.4 &62.9 &\textbf{86.7} &72.7  \\ \hline
     IDAL w/o FPN &\underline{58.6} &77.2  &80.1 &69.2 &76.4 &76.3   &\underline{70.8}  &56.9 &\underline{82.4}  & \underline{77.6}  &63.6 &84.2 &\underline{72.8} \\ \hline
     IDAL &\textbf{59.8}        &\underline{77.8}  &80.8 &\textbf{69.8} &76.9 &77.0    &\textbf{71.6}  &\underline{57.4} &\textbf{82.9}  & \textbf{78.5}  &\underline{64.1} &85.6 &\textbf{73.5}\\ \hline
\end{tabular}
}
\end{center}
\end{table*}

\begin{table*}
\caption{Accuracy(\%) on DomainNet for UDA (ResNet-101). In each sub-table, the column-wise domains are selected as the source domain and the row-wise domains are selected as the target domain. Highest accuracy is marked as bold and second highest is underline.}\label{dom}
\vspace{2.5mm}
  \setlength{\abovecaptionskip}{0.cm}
  \setlength{\belowcaptionskip}{0.cm}
  
 \centering
 \resizebox{\textwidth}{!}{
 \setlength{\tabcolsep}{0.5mm}{
   \begin{tabular}{|c|ccccccc||c|ccccccc||c|ccccccc|}
   \hline
   \textbf{ADDA} \cite{DBLP:journals/corr/TzengHSD17}  & clp   & inf   & pnt   & qdr   & rel   & skt   & Avg. & \textbf{DANN} \cite{ganin2016domain}  & clp   & inf   & pnt   & qdr   & rel   & skt   & Avg.  & \textbf{MIMTFL} \cite{10.1007/978-3-030-58592-1_35}  & clp   & inf   & pnt   & qdr   & rel   & skt   & Avg.  \\
   \hline
   \hline
   clp   & -     & 11.2  & 24.1  & 3.2   & 41.9  & 30.7  & 22.2  & clp   & -    & 15.5 & 34.8 & 9.5  & 50.8 & 41.4 & 30.4   & clp   & -    & 15.1 & 35.6 & 10.7 & 51.5 & 43.1 & 31.2  \\
   inf   & 19.1  & -     & 16.4  & 3.2   & 26.9  & 14.6  & 16.0  & inf   & 31.8 & -    & 30.2 & 3.8  & 44.8 & 25.7 & 27.3   & inf   & 32.1 & -    & 31.0 & 2.9  & 48.5 & \underline{31.0} & 29.1  \\
   pnt   & 31.2  & 9.5   & -     & \textbf{8.4}   & 39.1  & 25.4  & 22.7  & pnt   & 39.6 & 15.1 & -    & 5.5  & 54.6 & 35.1 & 30.0   & pnt   & 40.1 & 14.7 & -    & 4.2  & 55.4 & 36.8 & 30.2  \\
   qdr   & 15.7  & 2.6   & 5.4   & -     & 9.9   & 11.9  & 9.1   & qdr   & 11.8 & 2.0  & 4.4  & -    & 9.8  & 8.4  & 7.3    & qdr   & 18.8 & 3.1  & 5.0  & -    & 16.0 & 13.8 & 11.3  \\
   rel   & 39.5  & 14.5  & 29.1  & \textbf{12.1}  & -     & 25.7  & 24.2  & rel   & 47.5 & 17.9 & 47.0 & 6.3  & -    & 37.3 & 31.2   & rel   & 48.5 & 19.0 & 47.6 & 5.8  & -    & 39.4 & 32.1  \\
   skt   & 35.3  & 8.9   & 25.2  & \underline{14.9}  & 37.6  & -     & 25.4  & skt   & 47.9 & 13.9 & 34.5 & 10.4 & 46.8 & -    & 30.7   & skt   & 51.7 & 16.5 & 40.3 & 12.3 & 53.5 & -    & 34.9  \\
   Avg.  & 28.2  & 9.3   & 20.1  & \underline{8.4}   & 31.1  & 21.7  & 19.8  & Avg.  & 35.7 & 12.9 & 30.2 & 7.1  & 41.4 & 29.6 & 26.1   & Avg.  & 38.2 & 13.7 & 31.9 & 7.2  & 45.0 & 32.8 & 28.1  \\
   \hline
   \hline
   \textbf{ResNet-101} \cite{jian2016deep} & clp   & inf   & pnt   & qdr   & rel   & skt   & Avg.  & \textbf{CDAN$^{\dagger}$} \cite{long2018conditional}  & clp   & inf   & pnt   & qdr   & rel   & skt   & Avg.  &  \textbf{MDD$^{\dagger}$} \cite{pmlr-v97-zhang19i} & clp   & inf   & pnt   & qdr   & rel   & skt   & Avg.  \\
   \hline
   \hline
   clp   &  -   & 19.3  & 37.5  & \underline{11.1}  & 52.2  & 41.0  & 32.2  & clp  &  -   & 20.4 & 36.6 & 9.0  & 50.7 & 42.3 & 31.8  & clp   & -    & \underline{20.5} & 40.7 & 6.2  & 52.5 & 42.1 & 32.4 \\
   inf   & 30.2 & -     & 31.2  & 3.6   & 44.0  & 27.9  & 27.4  & inf  & 27.5 &  -   & 25.7 & 1.8  & 34.7 & 20.1 & 22.0  & inf   & \underline{33.0} & -    & 33.8 & 2.6  & 46.2 & 24.5 & 28.0 \\
   pnt   & 39.6 & 18.7  & -     & 4.9   & 54.5  & 36.3  & 30.8  & pnt  & 42.6 & 20.0 &  -   & 2.5  & 55.6 & 38.5 & 31.8  & pnt   & 43.7 & \underline{20.4} & -    & 2.8  & 51.2 & 41.7 & 32.0 \\
   qdr   & 7.0  & 0.9   & 1.4   & -     & 4.1   & 8.3   & 4.3   & qdr  & 21.0 & 4.5  & 8.1  &  -   & 14.3 & 15.7 & 12.7  & qdr   & 18.4 & 3.0  & 8.1  & -    & 12.9 & 11.8 & 10.8 \\
   rel   & 48.4 & 22.2  & 49.4  & 6.4   & -     & 38.8  & 33.0  & rel  & 51.9 & 23.3 & 50.4 & 5.4  &  -   & 41.4 & 34.5  & rel   & 52.8 & 21.6 & 47.8 & 4.2  & -    & 41.2 & 33.5 \\
   skt   & 46.9 & 15.4  & 37.0  & 10.9  & 47.0  & -     & 31.4  & skt  & 50.8 & 20.3 & 43.0 & 2.9  & 50.8 &  -   & 33.6  & skt   & 54.3 & 17.5 & 43.1 & 5.7  & 54.2 & -    & 35.0 \\
   Avg.  & 34.4 & 15.3  & 31.3  & 7.4   & 40.4  & 30.5  & 26.6  & Avg. & 38.8 & 17.7 & 32.8 & 4.3  & 41.2 & 31.6 & 27.7  & Avg.  & 40.4 & 16.6 & 34.7 & 4.3  & 43.4 & 32.3 & 28.6 \\
   \hline
   \hline
   \textbf{SCDA} \cite{DBLP:journals/corr/abs-2108-05720} & clp   & inf   & pnt   & qdr   & rel   & skt   & Avg.  & \textbf{CDAN + SCDA } \cite{DBLP:journals/corr/abs-2108-05720}  & clp   & inf   & pnt   & qdr   & rel   & skt   & Avg.  &  \textbf{MDD + SCDA} \cite{DBLP:journals/corr/abs-2108-05720} & clp   & inf   & pnt   & qdr   & rel   & skt   & Avg.  \\
   \hline
   \hline
   clp   &  -   & 18.6 & 39.3 & 5.1  & 55.0 & 44.1 & 32.4     & clp  &  -   & 19.5 & \underline{40.4} & 10.3 & \underline{56.7} & \underline{46.0} & \underline{34.6}  & clp   & -    & 20.4 & \textbf{43.3} & \textbf{15.2} & \textbf{59.3} & \textbf{46.5} & \textbf{36.9} \\
   inf   & 29.6 &   -  & 34.0 & 1.4  & 46.3 & 25.4 & 27.3     & inf  & \textbf{35.6} &  -   & \textbf{36.7} & \underline{4.5}  & \textbf{50.3} & 29.9 & \textbf{31.4}  & inf   & 32.7 & -    & \underline{34.5} & \textbf{6.3}  & \underline{47.6} & 29.2 & \underline{30.1} \\
   pnt   & 44.1 & 19.0 &   -  & 2.6  & 56.2 & \underline{42.0} & 32.8     & pnt  & 45.6 & 20.0 &  -   & 4.2  & \underline{56.8} & 41.9 & 33.7  & pnt   & \underline{46.4} & 19.9 & -    & \underline{8.1 } & \textbf{58.8} & \textbf{42.9} & \underline{35.2}\\
   qdr   & 30.0 & 4.9  & \underline{15.0}  &  -  & \underline{25.4} & \underline{19.8} & \underline{19.0}     & qdr  & 28.3 & 4.8  & 11.5 &  -   & 20.9 & 19.2 & 17.0  & qdr   & \underline{31.1} & \textbf{6.6}  & \textbf{18.0} & -    & \textbf{28.8} & \textbf{22.0} & \textbf{21.3} \\
   rel   & 54.0 & 22.5 & 51.9 & 2.3  &   -  & \underline{42.5} & 34.6     & rel  & \textbf{55.5} & 22.8 & \underline{53.7} & 3.2  &  -   & 42.1 & \underline{35.5}  & rel   & \underline{55.5} & \underline{23.7} & 52.9 & \underline{9.5}  & -    & \textbf{45.2} & \textbf{37.4} \\
   skt   & 55.6 & 18.5 & 44.7 & 6.4  & 53.2 &  -   & 35.7     & skt  & \textbf{58.4} & \underline{21.1} & \textbf{47.8} & 10.6 & \underline{56.5} &  -   & \textbf{38.9}  & skt   & \underline{55.8} & 20.1 & 46.5 & \textbf{15.0} & \textbf{56.7} & -    & \underline{38.8} \\
   Avg.  & 42.6 & 16.7 & 37.0 & 3.6  & 47.2 & 34.8 & {30.3}    & Avg. & \underline{44.7} & 17.6 & \underline{38.0} & 6.6 & \underline{48.2} & \underline{35.8} & \underline{31.8}  & Avg.  & 44.3 & \underline{18.1} & \textbf{39.0} & \textbf{10.8}  & \textbf{50.2} & \textbf{37.2} & \textbf{33.3} \\
   \hline
   \end{tabular}
   }}
 \label{tab:domainnet}
 \vspace{2.5mm}
 
 \vspace{-1mm}
\end{table*}

\begin{table}[h]
\caption{Accuracy (\%) on DomainNet for UDA with IDAL(ours). Highest accuracy is marked as bold and second highest is underline.}\label{dom1}
\vspace{2.5mm}
\centering

\begin{tabular}{|c|c|c|c|c|c|c|c|}
\hline
 & clp & inf & pnt & qdr & rel & skt & Avg. \\
\hline
clp & - & \textbf{20.5} & 36.8 & 9.2 & 54.8 & 44.5 & 33.2 \\
\hline
inf & 32.4 & - & 25.8 & 2.5 & 31.2 & \textbf{41.2} & 26.6 \\
\hline
pnt & \textbf{72.5} & \textbf{25.0} & - & 4.5 & 52.7 & 37.6 & \textbf{38.5} \\
\hline
qdr & \textbf{47.4} & \underline{5.3} & 8.4 & - & 11.8 & 12.5 & 17.1 \\
\hline
rel & 51.5 & \textbf{25.2} & \textbf{54.5} & 2.9 & - & 34.8 & 33.8 \\
\hline
skt & 49.8 & \textbf{24.8} & \underline{47.4} & 12.8 & 51.6 & - & 37.3 \\
\hline
Avg. & \textbf{50.7} & \textbf{20.2} & 34.6 & 6.4 & 40.4 & 34.1 & {31.1} \\
\hline
\end{tabular}
\end{table}

\begin{table*}[h]
\caption{Comparison with SoTA methods on VisDA-2017. IDAL(ours) is reported with and without(w/o) FPN. The best performance is marked as bold, and the second best is underlined.}\label{t3}
\begin{center}
\resizebox{\textwidth}{!}{%
\begin{tabular}{|c|c|c|c|c|c|c|c|c|c|c|c|c|c|}
    \hline
    \textbf{Model} & \textbf{plane} & \textbf{bcycl} & \textbf{bus} & \textbf{car} & \textbf{horse} & \textbf{knife} & \textbf{mcycl} & \textbf{person} & \textbf{plant} & \textbf{sktbrd} & \textbf{train} & \textbf{truck} &\textbf{Avg.}\\ \hline \hline
     ResNet-50 \cite{jian2016deep} & 55.1   &53.3  &61.9  & 59.1  &80.6 &17.9 &79.7 &31.2  &81.0 &26.5 &73.5 &8.5  &52.4  \\ 
     BNM \cite{cui2020towards}  &	89.6	&	61.5	&	76.9	&	55.0	&	89.3	&	69.1	&	81.3	&	65.5	&	90.0	&	47.3	&	89.1	&	30.1	&	70.4	\\ 
     MCD \cite{saito2018maximum} &	87.0	&	60.9	&	83.7	&	64.0	&	88.9	&	79.6	&	84.7  &	76.9	&	88.6	&	40.3	&	83.0	&	25.8	&	71.9	\\ 
     SWD \cite{lee2019sliced} &	90.8	&	82.5	&	81.7	&	70.5	&	91.7	&	69.5	&	86.3	&	77.5	&	87.4	&	63.6	&	85.6	&	29.2	&	76.4	\\ 
     FixBi \cite{na2021fixbi} & \textbf{96.1}   &87.8  &\textbf{90.5}  & \textbf{90.3}  &\textbf{96.8} &95.3 &\textbf{92.8}  &\textbf{88.7} &\textbf{97.2} &\textbf{94.2} &\textbf{90.9} &25.7 &\underline{87.2}  \\ \hline
     IDAL w/o FPN  &94.1        &\underline{88.6}  &89.2 &78.7 &94.9 &\underline{98.2}    &88.5 &84.6 &94.7 & 90.3  &88.4 &\underline{51.3} &86.8\\ 
     IDAL  &\underline{94.7}        &\textbf{89.0}  &\underline{89.6 } &\underline{79.0} &\underline{95.6} &\textbf{98.7}    &\underline{89.4} &\underline{85.2} &\underline{95.6} & \underline{90.5}  & \underline{88.9} &\textbf{52.6} &\textbf{87.4}\\ \hline
\end{tabular}
}
\end{center}
\end{table*}

\begin{table*}[h]
\caption{Comparison with SoTA methods on Office-31. IDAL(ours) is reported with and without(w/o) FPN. The best performance is marked as bold, and the second best is underlined.}\label{t1}
\begin{center}
\begin{tabular}{|p{2.4cm}|p{1cm}|p{1.1cm}|p{1.1cm}|p{1.1cm}|p{1.1cm}|p{1.1cm}|p{1.2cm}|}
    \hline
    Method	&	A $\rightarrow$ D 	&	A $\rightarrow$ W	&	D $\rightarrow$ W	&	W $\rightarrow$ D	&	D $\rightarrow$ A	&	W $\rightarrow$ A	&	Avg \\ \hline
     ResNet-50 \cite{jian2016deep}	&	68.9	&	68.4	&	96.7	&	99.3	&	62.5	&	60.7	&	76.1 \\ 
     DANN \cite{ganin2016domain}	&	79.7	&	82.0	&	96.9	&	99.1	&	68.2	&	67.4	&	82.2 \\ 
     CDAN \cite{long2018conditional} &	92.9	&	94.1	&	98.6	&	100.0	&	71.0	&	69.3	&	87.7 \\ 
     MDD	\cite{pmlr-v97-zhang19i} &	93.5	&	94.5	&	98.4	&	100.0	&	74.6	&	72.2	&	88.9 \\ 
     GVB-GD \cite{cui2020gradually} &	95.0	&	94.8	&	98.7	&	100.0	&	73.4	&	73.7	&	89.3 \\ 
     SRDC  \cite{tang2020unsupervised} &	\textbf{95.8}	&	95.7	&	99.2	&	100.0	&76.7	&77.1	&90.8 \\ 
     SHOT	\cite{liang2020we} &	93.1	&	90.9	&	98.8 &	99.9	&	74.5	&	74.8	&	88.7 \\ 
     f-DAL \cite{acuna2021f}	&	94.8	&	93.4	&	99.0 &	100.0	&	73.6	&	74.6	&	89.2 \\ 
     FixBi \cite{na2021fixbi} &	95.0 &	\textbf{96.1}	&	\textbf{99.3} &	\underline{100.0}	&	\textbf{78.7}	&	\textbf{79.4}	&	\textbf{91.4} \\ \hline
     IDAL w/o FPN &	94.4	&	95.0	&	99.0 &	100.0	&	75.6	&	76.6	&	90.1 \\ 
      IDAL &	\underline{95.6} &	\underline{95.7}	&	\underline{99.1} &	\textbf{100.0}	&	\underline{77.3}	& \underline{77.1}	&	\underline{90.8} \\ \hline
\end{tabular}
\end{center}
\end{table*}

\section{Experiments and Results}

To validate the efficacy of our model, we undertake extensive investigations on well-established benchmarks and juxtapose our results with those achieved by state-of-the-art UDA methods. We also studied the impact of using a feature pyramid network (FPN)~\cite{lin2017feature} for domain adaptation for classification. Additionally, we studied how the feature (representation) space of the target domain evolves during training. We also studied the contribution of various components of the loss function. We also studied the convergence speed of our method compared to FixBi~\cite{na2021fixbi}.

\subsection{Datasets}

To evaluate the proposed method, we conducted experiments on  benchmark UDA datasets -- including Office-31~\cite{saenko2010adapting}, Office-Home~\cite{venkateswara2017deep}, VisDA-2017~\cite{pei2018multi}, and DomainNet~\cite{peng2019moment}. The details of the datasets and transfer tasks on these datasets are given below:

The Office-Home dataset is a key benchmark with 15,500 images across 65 classes and four domains: Artistic, Clip Art, Product, and Real-World, used to assess twelve transfer tasks. Office-31, another pivotal dataset, contains 4,110 images in 31 classes from Amazon, Webcam, and DSLR domains, evaluating six transfer tasks. VisDA-2017, aimed at cross-domain generalization, includes Synthetic and Real source domains with 12 categories, using the ImageNet validation set as the target. DomainNet, the largest dataset for domain adaptation, features about 0.6 million images across 345 categories from six domains (Clipart, Infograph, Painting, Quickdraw, Real, and Sketch), supporting 30 adaptation tasks, showcasing its scale and diversity in visual domain adaptation challenges.

\subsection{Implementation details}

All the experiments were conducted on an NVIDIA A100 in PyTorch, using the CNN-based neural network (ResNet-50) pre-trained on ImageNet~\cite{he2016deep} and feature pyramid network as the backbone for our proposed model. The base learning rate is 0.00001 with a batch size of 32, and we train models by 50 epochs. The hyper-parameters were $\beta$=0.05 , $\gamma$=0.1 , $\delta$=0.15 and $\eta$=0.15 for the experiment of Office-31 dataset. Similarly for Office-Home, the hyper-parameters were $\beta$=0.05 , $\gamma$=0.21 , $\delta$=0.25 and $\eta$=0.25 , for the VisDA-2017 dataset the hyper parameters were $\beta$=0.05 , $\gamma$=0.3 , $\delta$=0.25 and $\eta$=0.25 ,and for the Domain-Net dataset the hyper parameters were $\beta$=0.05 , $\gamma$=0.01 , $\delta$=0.2 and $\eta$=0.25.We have used AdamW~\cite{loshchilov2018decoupled} with a momentum of 0.9, and a weight decay of 0.001 as the optimizer. We adhere to the standard procedure for unsupervised domain adaptation (UDA), wherein we make use of both labeled source samples and unlabeled target samples during the training process.  For a fair comparison with prior works, we also conduct experiments with the same backbones such as ResNet-50~\cite{jian2016deep}, DANN~\cite{ganin2016domain}, CDAN~\cite{long2018conditional}, MDD~\cite{pmlr-v97-zhang19i}, GVB-GD~\cite{cui2020gradually}, SRDC~\cite{tang2020unsupervised}, FixBi~\cite{na2021fixbi}, SHOT~\cite{liang2020we}, SDAT~\cite{rangwani2022closer}, f-DAL~\cite{acuna2021f}, BNM~\cite{cui2020towards}, MCD~\cite{saito2018maximum} and SWD~\cite{lee2019sliced} for demonstration of results with different-different datasets.

\subsection{UDA benchmarks}
We assess the performance of our proposed model by conducting comparisons with state-of-the-art methods that rely on ResNet-based architectures. In these experiments, we employ ResNet-50 as the underlying architecture for our evaluations across the Office-Home, Office-31 and VisDA-2017 datasets and ResNet-101 for the DomainNet dataset. Importantly, each ResNet-50 and ResNet-101 backbone is trained exclusively on source data and subsequently subjected to testing using target data.

\begin{figure*}[!]
\centering
\includegraphics[height=4cm,width=12cm]{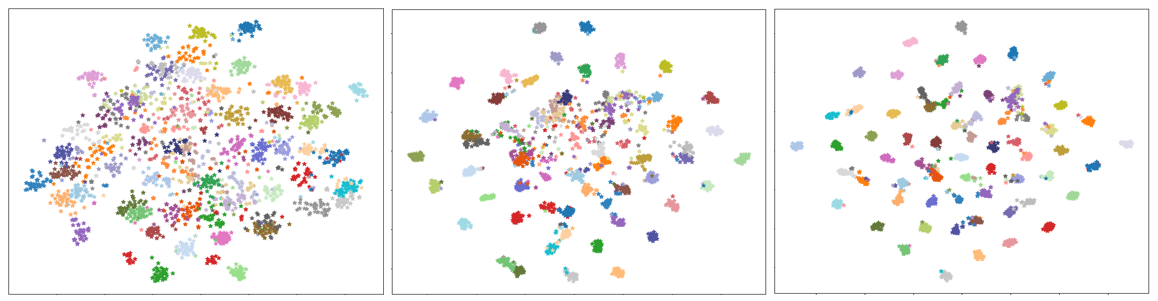}
\caption{Two-dimensional visual representation, generated using t-SNE~\cite{van2008visualizing}, illustrates the evolution of the domain adaptation task from R to P on the Office-Home dataset. The images depict untrained \emph{target} samples (leftmost), training progress after 4 epochs (middle), and the state after 9 epochs (rightmost). Initially, overlapping classes gradually reconfigure into distinct clusters during the training process.}
\label{tsne}
\end{figure*}

\begin{table*}[h]
\caption{Performance comparison of IDAL with different combinations of losses on Office-Home. The best performance is marked as bold.}\label{t4}
\begin{center}
\resizebox{\textwidth}{!}{%
\begin{tabular}{|c|c|c|c|c|c|c|c|c|c|c|c|c|c|c|c|c|c|} 
    \hline
    \textbf{M1} & 
    \textbf{MMD} & 
    \textbf{MCC} & 
    \textbf{PLMMD} & 
    \textbf{A$\rightarrow$C} & \textbf{A$\rightarrow$P} & \textbf{A$\rightarrow$R} & \textbf{C$\rightarrow$A} & \textbf{C$\rightarrow$P} & \textbf{C$\rightarrow$R} & \textbf{P$\rightarrow$A} & \textbf{P$\rightarrow$C} & \textbf{P$\rightarrow$R} & \textbf{R$\rightarrow$A} & \textbf{R$\rightarrow$C} & \textbf{R$\rightarrow$P} &\textbf{Avg.}\\ \hline \hline
     $\checkmark$  & $\times$ & $\times$ & $\times$ & 50.7   &70.6  &76.0  & 57.6  &70.0 &70.0 &57.4  &50.9 &77.3 &70.9 &56.7 &81.6 &65.8  \\ 
     \checkmark  & \checkmark & $\times$ & $\times$ &	59.4	&	76.8	&	80.3	&	69.1	&	75.7	&	76.2	&	69.7	&	56.8	&	82.3	&	78.4	&	63.4	&	84.6	&	72.7	\\  
     \checkmark  & \checkmark & \checkmark & $\times$ &	59.6	&	77.6	&	80.4	&	69.3	&	76.0	&	76.8	&	71.2	&	57.1	&	82.7	&	78.5	&	64.0	&	85.2	&	73.2	\\   \hline   
     \checkmark & \checkmark & \checkmark & \checkmark &\textbf{59.8}        &\textbf{77.8}  &\textbf{80.8} &\textbf{69.8} &\textbf{76.9} &\textbf{77.0}    &\textbf{71.6}  &\textbf{57.4} &\textbf{82.9}  & \textbf{78.5}  &\textbf{64.1} &\textbf{85.6} &\textbf{73.5}\\ \hline
\end{tabular}
}
\end{center}
\end{table*}

Table~\ref{t2} presents quantitative results with various backbones, demonstrating our proposed model's consistent superiority over the state-of-the-art, specifically achieving an impressive average accuracy improvement of over 0.8\% when compared to the FixBi model for the Office-Home dataset. Table~\ref{t3} further showcases our model's superiority, surpassing the current state-of-the-art and attaining a notable 0.2\% average accuracy improvement over the FixBi model for the VisDA-2017 dataset. Table~\ref{t1} illustrates results using diverse backbones, highlighting our model's performance, which is comparable with the current state-of-the-art for the Office-31 dataset. Table~\ref{dom} and \ref{dom1} depict the performance of our model with ResNet-101 + FPN as the feature extractor for the most challenging dataset of domain adaptation with comparable performance to state-of-the-art.

ResNet-50 combined with Feature Pyramid Network (FPN) for feature extraction offers multi-scale feature capture, superior object detection, and effective feature fusion. This versatile pairing, proven in various computer vision tasks, balances depth and scale, enhancing overall performance while reducing computational costs. Tables~\ref{t2},~\ref{t3}, and~\ref{t1} show the impact of having FPN with ResNet for feature extraction. We conducted ablation studies to understand the impact of the different feature extractors such as ConvMixer~\cite{trockman2022patches} and ResNet-101~\cite{jian2016deep}. However, the performance in these cases was worse than our reported results. We also compare the ResNet-based backbone and transformer-based backbone and notice a huge gap in parameter requirements. ResNet-based backbone needs relatively very less parameters compared to transformer-based backbones.

The tSNE~\cite{van2008visualizing} plot is shown in Figure~\ref{tsne} for the office-Home dataset for the task of domain adaptation when R is the source and P is the target. We can see the rapid evolution of a multi-modal distribution of the target domain features where classes (denoted by separate colors) get separated.

\subsection{Impact of loss components}
To gauge the influence of individual loss functions and their collective impact, we conducted a thorough experimental analysis. Our findings revealed that Minimum Class Confusion (MCC) loss functions enhance classification models by reducing class confusion, especially in scenarios with imbalanced class distributions. Concurrently, we observed that information maximization losses assist the classifier in prioritizing the most confidently aligned samples for domain adaptation. Additionally, the Maximum Mean Discrepancy (MMD) loss effectively narrows the gap between the mean embeddings of the two distributions. Table~\ref{t4} shows the effect of the individual loss function on the performance of our model IDAL for the Office-Home dataset and it indicates that our model IDAL performs best with a tailored combination of loss functions. By artfully combining these distinctive loss functions, we not only surpass the current state-of-the-art but also achieve a comprehensive solution that advances the field of classification models in diverse scenarios.




\section{Conclusions and Future Directions}

We proposed a novel method for unsupervised domain adaptation for image classification. We proposed a novel neural network architecture and a loss function. Architecturally, we have demonstrated that synergy between two deep learning architectures -- ResNet~\cite{he2016deep} and feature pyramidal network (FPN)~\cite{lin2017feature} -- complement each other to extract multi-scale features and effectively separate style (domain) and content (class) information components. Our ablation studies confirm the importance of using FPN with ResNet. The proposed loss component PLMMD and judiciously chosen existing loss components leads to significant improvements in unsupervised domain adaptation (UDA) performance that can surpass the performance of CNNs using other UDA methods. Our ablation study confirmed the importance of each of the loss components. Additionally, using the proposed loss led to faster convergence and a rapid evolution of a class-wise multi-modal distribution of the target domain features. 

In the future, computationally heavier architectures, such as, vision transformers~\cite{dosovitskiy2020image} and its derivatives may be used for further improvements in domain adaptation. Additionally, the proposed loss function may be adapted for other tasks, such as semantic segmentation and object detection.
{\small
\bibliographystyle{ieee_fullname}
\bibliography{egbib}
}

\end{document}